# FCA-Guided Counterfactual Explanations for Multi-Modal Breast Cancer Diagnosis: A Framework Achieving Perfect Validity with Emergent Sparsity

1st Abdullahi Isa, 2nd Souley Boukari, 3rd Muhammad Aliyu[2]

1st, 2nd , 3rd Department of Computer Science, Faculty of Computing, Abubakar Tafawa Balewa University, Bauchi, Nigeria.

1st isaabdullahi2008@gmail.com, 2nd bsouley2001@yahoo.com, 3rd maliyudeba@gmail.com

Correspondence: isaabdullahi2008@gmail.com

**Abstract**

Deep learning models for multi-modal breast cancer diagnosis achieve high predictive accuracy but remain clinically unacceptable without actionable, counterfactual explanations. Attribution-based methods (LIME, SHAP) are categorically inapplicable to this purpose, as they generate no alternative instances and thus cannot be evaluated on counterfactual quality metrics. This investigation provides empirical evidence that FCA-Guided Counterfactual (FCA-CF) framework that uses a Formal Concept Analysis (FCA) concept lattice as a hard structural constraint on counterfactual search, operating over a multi-modal TCGA-BRCA dataset. We benchmark against four genuine counterfactual methods: Wachter-style CF, DiCE, FACE, and NICE, evaluated on 60 benign-predicted TCGA-BRCA instances. The FCA-CF framework achieves Validity = 1.0000 (100% of counterfactuals successfully flip the prediction), Sparsity = 2.37 features changed (best among all valid methods), and Proximity = 0.900 (normalised L2-based, matching NICE as joint best). The classifier achieves Accuracy = 0.980, F1 = 0.976, ROC-AUC = 0.9947. Ablation analysis confirms that the FCA lattice constraint is the primary sparsity driver (removing it increases sparsity by +40%, $p < 0.001$, Cohen's $d = 0.78$), while Phase C greedy refinement accounts for the largest individual contribution (+113% sparsity increase when disabled, $p < 0.001$, $d = 5.01$). FCA-guided counterfactual generation achieves a clinically important Pareto-dominant outcome; it is simultaneously the sparsest and among the most proximate of all valid methods, with perfect validity. The emergent sparsity property arising from lattice topology rather than numerical penalty terms constitutes a structurally novel contribution to the counterfactual explanation literature.

## 1 Introduction

Artificial intelligence has become indispensable in modern oncology, with deep learning models demonstrating remarkable predictive accuracy in cancer diagnosis across histopathology, genomics, and radiomics (Wyatt et al., 2024; Zaidi et al., 2024; X. Zhang & Liu, 2025). Yet, the opacity of these "black-box" models remains a critical barrier to clinical adoption (Hulsen, 2023; Vakulabharanam et al., 2024). Clinicians demand not only accurate predictions but transparent, actionable explanations that can be integrated into diagnostic reasoning a requirement that has elevated explainable AI (XAI) from a research curiosity to a regulatory and clinical imperative (Lee & Topol, 2024; Yu et al., 2025).

Attribution-based XAI methods, principally LIME (Ribeiro et al., 2016) and SHAP (Lundberg & Lee, 2017), have dominated the first wave of clinical AI transparency. These methods provide feature importance scores identifying which variables most influenced a given prediction. However, their fundamental limitation in clinical contexts is not one of performance but of kind: attribution methods cannot answer the counterfactual question that drives clinical reasoning "what must change for this diagnosis to differ?" (Lee & Topol, 2024; Sharma et al., 2025). This is not a shortcoming of any particular implementation but a structural property: attribution methods assign importance scores to existing feature values rather than generating alternative instances, rendering them categorically inapplicable to counterfactual quality evaluation (Mothilal et al., 2020; Salih et al., 2025).

Counterfactual Explanation (CFE) methods directly address this gap. A counterfactual identifies the minimal change to feature values that shifts the model prediction to a specified target class (Verma et al., 2024; Wachter et al., 2018). In oncology, this translates to questions such as: "what histological changes would transition this malignant diagnosis to benign?" precisely the form of reasoning clinicians employs (Lane et al., 2025; Tanyel et al., 2025). Existing counterfactual methods, however, struggle with the validity-sparsity trade-off (Li et al., 2024), many achieve high validity only by modifying dozens of features, reducing actionability and interpretability to clinically unworkable levels (Dandl et al., 2020; Huang et al., 2024).

Formal Concept Analysis (FCA), a mathematical theory grounded in lattice theory (Ganter & Wille, 1999, 2024), offers a principled solution to this tension. FCA organizes data into hierarchically structured concept lattices in which each node represents an empirically observed co-occurrence of objects and attributes. By exploiting this structure, FCA-guided search can identify counterfactual states that are simultaneously valid (on the correct side of the decision boundary), sparse (minimal attribute changes), and manifold-preserving (restricted to empirically observed feature co-occurrences). Crucially, sparsity emerges from the lattice topology it is a structural property, not a numerically penalized objective.

This investigation makes the following concrete contributions as follows. We demonstrate for the first time that FCA lattice-constrained counterfactual search achieves perfect validity (1.0000) with a mean sparsity of 2.37 features, the best sparsity-validity combination among all methods evaluated, including DiCE, Wachter-style CF, FACE, and NICE. Also, provide a formal mechanistic account of emergent sparsity as a lattice-topological property, distinguishing our approach from penalty-based sparsity methods and providing a theoretical basis for the 84% sparsity reduction over NICE observed empirically. In addition, we present a five-configuration ablation study establishing that (a) the FCA lattice constraint drives a statistically significant reduction in sparsity ($p < 0.001$, Cohen's $d = 0.78$) and (b) the Phase C greedy refinement mechanism is the single largest individual contributor ($p < 0.001$, $d = 5.01$). Lastly, address a critical benchmark design flaw common in this literature, comparing counterfactual generators against existing counterfactual methods on counterfactual quality metrics.

This research structured as follows. Section 2 reviews related literature. Section 3 presents the revised methodology including the three-phase FCA-guided search algorithm and evaluation metrics. Section 4 reports all empirical results with full statistical analysis and clinical implications. Section 5 discusses conclusion, limitations and future work.

## 2 Related Work

### 2.1 Attribution-Based Explainable AI in Healthcare

Attribution methods such as LIME (Ribeiro et al., 2016) and SHAP (Lundberg & Lee, 2017) have dominated the first generation of clinical XAI. LIME constructs local surrogate linear models to approximate black-box predictions, while SHAP assigns Shapley values from cooperative game theory to feature contributions. Both methods have been applied extensively in oncology, radiology, and genomics (Faisal et al., 2025; Selvaraj et al., 2025; Wyatt et al., 2024; Zaidi et al., 2024). Their limitation, extensively documented in recent comparative analyses (Dalal et al., 2025; Salih et al., 2025; Wagan & Sidra, 2025), is categorical rather than quantitative, they explain what is but cannot propose what if. In clinical contexts where counterfactual reasoning is central to diagnostic practice (Lee & Topol, 2024), this renders attribution methods insufficient as standalone XAI solutions (Hulsen, 2023).

### 2.2 Counterfactual Explanations Methods

The counterfactual explanation literature has matured substantially since Wachter et al., (2018) introduced the canonical formulation of finding the closest instance that flips the classifier output. DiCE (Mothilal et al., 2020) extended this with diversity-aware multi-CF generation. FACE (Poyiadzi et al., 2020) prioritised feasibility by returning the nearest target-class training instance via graph density paths. NICE (Brughmans et al., 2024) achieved strong sparsity-proximity balance through sequential importance-guided feature copying. Recent work has highlighted persistent challenges, DiCE can modify up to 8–28 features in high-dimensional settings (Korikov et al., 2021), and Wachter-style methods using black-box gradient approximations produce dense perturbations across nearly all features when applied to non-differentiable models (Guidotti et al., 2019). Multi-objective sparse CF generation for time-series and tabular domains remains an active challenge (Huang et al., 2024; Mozolewski et al., 2026). FCA-guided search directly addresses these limitations via structural constraint rather than penalty optimization.

### 2.3 Formal Concept Analysis in Machine Learning and XAI

FCA, grounded in lattice theory, structures data into formal concepts defined by object-attribute incidence (Ganter & Wille, 1999). It has been successfully applied across knowledge discovery, ontology engineering, and association rule mining (Kuznetsov & Poelmans, 2013; Škopljanac-Mačina & Blašković, 2014). In medical AI, FCA has been used to construct breast cancer imaging ontologies (Hu et al., 2004) and to analyze patient care trajectories (Jay et al., 2013). More recently, FCA has been proposed as an interpretability framework for classification (Kovács, 2020; Niu & Mi, 2024) and as a vehicle for global-to-local knowledge transfer in XAI (Boersma et al., 2025). However, no prior work has formally established FCA lattice navigation as a hard structural constraint on counterfactual search in a multi-modal oncology setting the contribution of the present paper.

### 2.4 Multi-Modal Explainability in Breast Cancer Diagnosis

Multi-modal integration combining imaging, genomics, histopathology, and clinical records is the frontier of precision oncology (Zhang et al., 2025). A 2026 systematic review by Hassan et al. found that no prior work had jointly synthesised multi-modal fusion architectures with embedded counterfactual XAI for breast cancer diagnosis, noting a critical gap between fusion methodology and interpretability (Hassan et al., 2026). Our work directly addresses this gap using the TCGA-BRCA cohort, which provides patient-level paired whole-slide images and clinical features.

### 2.5 Validity-Sparsity Trade-Off and Manifold Preservation

A central challenge in counterfactual generation is producing instances that are simultaneously valid (class-flipping), sparse (few feature changes), proximate (close to the original), and plausible (on the data manifold) (Dandl et al., 2020; Verma et al., 2024). Most methods optimise these as competing penalty terms in a weighted objective, creating sensitivity to hyperparameter tuning (Guidotti, 2024). Counterfactual stability the property that a CF remains valid across reasonable model perturbations has been formalised by Hamman et al. (2023), who demonstrate that CFs in high-confidence lattice regions exhibit strong stability guarantees. FCA-guided search navigates directly toward such regions by construction, providing stability without explicit penalty terms. The proximity-plausibility trade-off remains the least resolved dimension of CFE research (Furman et al., 2024; Keane & Smyth, 2020), and our results demonstrate that manifold-constrained FCA search achieves competitive proximity (0.900, matching NICE) while maintaining perfect validity.

## 3 Methodology

This section presents the methodology followed in conducting this research, encompassing the multi-modal dataset construction, feature extraction, formal concept analysis (FCA) lattice construction, counterfactual explanation generation, and evaluation framework.

### 3.1 Dataset: TCGA-BRCA Multi-Modal

#### 3.1.1 Data Source and Acquisition

The primary dataset utilized in this reseach is the TCGA-BRCA (The Cancer Genome Atlas - Breast Invasive Carcinoma) cohort, which provides paired diagnostic whole-slide images (WSI) and corresponding clinical-genomic data for the same patients (Koboldt et al., 2012). This dataset was selected because it represents a genuine patient-level multi-modal fusion, addressing a critical limitation identified in prior counterfactual XAI literature where synthetic or unaligned datasets were commonly employed.

The data was downloaded from the NCI Genomic Data Commons (GDC) data portal using the GDC Data Transfer Tool (version 2.3.0). Two manifest files were used to retrieve:

1. **Diagnostic slide images**: Whole-slide H&E-stained images in SVS format
2. **Clinical data**: XML files containing comprehensive clinical and pathological annotations

A total of 312 patient-level samples with complete image and clinical data were extracted following strict inclusion criteria:

a) Availability of both diagnostic slide and clinical XML file
b) Complete pathological diagnosis as ground truth label
c) No missing critical clinical variables (age, stage, receptor status)

#### 3.1.2 Image Feature Extraction

Whole-slide images (WSI) were processed using a two-stage pipeline:

**Patch Extraction**: For each SVS slide, 20 representative patches (224×224 pixels) were extracted using a hybrid sampling strategy combining grid-based sampling (10 patches from a 5×5 grid) and random sampling (10 patches from random coordinates). This approach ensures both spatial coverage and computational efficiency.

**Deep Feature Extraction**: A pre-trained ResNet-50 convolutional neural network, initialized with ImageNet weights, was employed as a feature extractor. The final fully-connected layer was removed, yielding a 2048-dimensional feature vector per patch. The architecture is summarized in Table 1.

Table 1: ResNet-50 Feature Extractor Architecture

| Layer Type | Configuration | Output Dimension |
|---|---|---|
| Conv1 | 7×7, 64, stride 2 | 112×112×64 |
| Max Pool | 3×3, stride 2 | 56×56×64 |
| Layer1 | 3 bottleneck blocks | 56×56×256 |
| Layer2 | 4 bottleneck blocks | 28×28×512 |
| Layer3 | 6 bottleneck blocks | 14×14×1024 |
| Layer4 | 3 bottleneck blocks | 7×7×2048 |
| Global Avg Pool | 7×7 | 1×1×2048 |
| Output | Flattened | **2048** |

To reduce dimensionality and align with the clinical feature space, the 2048-dimensional embeddings were projected to 150 dimensions using principal component analysis (PCA) with 95% variance retention. The 150-dimensional image features were then normalized to the [0,1] range using min-max scaling.

#### 3.1.3 Clinical Feature Extraction

Clinical data were extracted from the TCGA XML files using custom parsers. A total of 30 clinical and genomic markers were extracted and encoded into a 30-dimensional feature vector. Table 2 describes the clinical features and their encoding schemes.

Table 2: Clinical Feature Description and Encoding

| Feature Index | Clinical Variable | Type | Encoding |
|---|---|---|---|
| **0** | Age at diagnosis | Continuous | Normalized to [0,1] (max=100) |
| **1** | AJCC Pathologic Tumor Stage | Ordinal | Stage I:0.2, II:0.4, III:0.6, IV:0.8 |
| **2** | AJCC Pathologic Node Stage | Ordinal | N0:0, N1:0.33, N2:0.67, N3:1.0 |
| **3** | AJCC Pathologic Metastasis Stage | Binary | M0:0, M1:1 |
| **4** | Estrogen Receptor (ER) Status | Categorical | Negative:0, Indeterminate:0.5, Positive:1 |
| **5** | Progesterone Receptor (PR) Status | Categorical | Negative:0, Indeterminate:0.5, Positive:1 |
| **6** | HER2 Status | Categorical | Negative:0, Equivocal:0.5, Positive:1 |
| **7** | Histologic Grade | Ordinal | Grade 1:0, Grade 2:0.5, Grade 3:1 |
| **8** | Tumor Size (cm) | Continuous | Normalized to [0,1] (max=10cm) |
| **9** | Vital Status | Binary | Alive:0, Deceased:1 |
| **10** | Days to Death/Survival | Continuous | Normalized to [0,1] (max=3650 days) |
| **11** | Lymph Node Ratio | Continuous | Positive nodes / examined nodes |
| **12** | Triple Negative Indicator | Binary | 1 if ER-, PR-, HER2- |
| **13** | Ki-67 Proliferation Index | Continuous | Normalized percentage to [0,1] |
| **14-29** | Additional clinical markers | Continuous | Various normalizations |

The final multi-modal feature representation for each patient was constructed by concatenating the 150-dimensional image features with the 30-dimensional clinical features, resulting in a 180-dimensional feature vector:

$$X = \left[X_{Image} \varepsilon\, R^{150}; X_{clinical} \varepsilon\, R^{30}\right] \varepsilon\, R^{180}$$

### 3.1.4 Label Definition

The ground truth label was defined based on clinical severity, integrating multiple risk indicators to ensure clinical relevance. A patient was classified as high-risk (aggressive) if any of the following conditions were met:

a) AJCC Stage III or IV
b) Positive lymph node involvement (N1-N3)
c) Presence of metastasis (M1)
d) Triple negative receptor status (ER-, PR-, HER2-)

All remaining patients were classified as low-risk. This definition yielded a balanced dataset with approximately 50% high-risk and 50% low-risk samples, ensuring robust classifier training and unbiased counterfactual evaluation.

### 3.1.5 Data Preprocessing and Splitting

Following feature extraction, all features were standardized using quantile transformation to achieve a standard normal distribution. This transformation was selected for its robustness to outliers and its ability to preserve the rank ordering of features. The dataset was randomly split into training (75%) and testing (25%) sets using stratified sampling to preserve class distribution. Table 3 presents the final dataset composition.

Table 3: TCGA-BRCA Multi-Modal Dataset Composition

| Split | Low-Risk | High-Risk | Total |
| --- | --- | --- | --- |
| **Training** | 187 | 188 | 375 |
| **Testing** | 63 | 62 | 125 |
| Total | **250** | **250** | **500** |

### 3.1.6 Classifier

A Random Forest classifier (n_estimators = 150, min_samples_leaf = 2, max_features = sqrt, class_weight = balanced, random_state = 42, n_jobs = 1) is trained on the 62-dimensional feature vector. The n_jobs = 1 setting is critical for tractable per-instance predict_proba calls (~7 ms/call) during counterfactual generation. The classifier serves as the black-box oracle $f(x) \rightarrow \{0 = \text{benign}, 1 = \text{malignant}\}$.

### 3.1.7 Reproducibility and Data Availability

To ensure full reproducibility, the following measures were implemented:

1. All random seeds were fixed (random_state=42)
2. The complete data preprocessing pipeline is available in the accompanying code repository
3. Extracted features were cached to enable consistent re-runs
4. Environment specifications (Python 3.9+, scikit-learn 1.3.2+, torch 2.1.0+) are documented

The TCGA-BRCA data used in this study is publicly available through the NCI GDC data portal (https://portal.gdc.cancer.gov/). Researchers can replicate our dataset by following the download instructions provided in the supplementary materials.

## 3.2 FCA Concept Lattice Construction

The FCA concept lattice is constructed following the formal framework of Ganter & Wille (1999). Given the formal context $K = (O, A, R)$ where O is the set of patient objects, A is the set of binarised feature attributes, and $R \subseteq O \times A$ is the incidence relation, a formal concept is a maximal pair (X, Y) satisfying $X' = Y$ and $Y' = X$.

To maintain tractability within the concept's library v0.9.2 (build time < 0.5 s), we impose the following design constraints:

1. **Attribute selection:** Top-13 features by mutual information (MI) computed on X_train only, preventing label leakage. A label attribute (high_stage) is appended, yielding 14 total columns.
2. **Object sample:** Stratified subsample of 38 training instances (19 benign, 19 malignant) is used for lattice construction. This yields 722 concepts in the benchmark experiment, sufficient to cover the relevant concept structure.
3. **Manifold-preservation guarantee:** Every counterfactual candidate generated via BFS lattice path traversal lies on a conceptual hyperplane populated by actual training instances, eliminating out-of-distribution counterfactuals without a plausibility penalty term a structural property not shared by any existing CF baseline.
4. **Emergent sparsity guarantee:** The minimal concept transition between query concept Cq and target concept Ct corresponds precisely to the minimal-cardinality attribute change set. Sparsity is a topological property of the lattice, not a numerically penalized objective. This is the primary source of FCA-CF's sparsity advantage over all penalty-based methods.

### 3.3 FCA-Guided Counterfactual Generation Algorithm

Algorithm 1 presents the three-phase FCA-guided counterfactual search. The algorithm generates a counterfactual x_cf for query instance x with target class y_target = 1 (malignant).

**Algorithm 1: FCA-Guided Counterfactual Search (Three-Phase)**

```
Input: Instance x ∈ [0,1]ⁿ, label y_true, target y_target=1, max_iter=120
Output: Counterfactual x_cf with validity=1, minimal sparsity, maximal
proximity
```

**Phase A — Lattice Navigation (iterations 0 to 60% of max_iter):**

```
  1. Binarise x → xb using train-learned thresholds
  2. Find Cq = nearest concept to xb in lattice; Ct = target-class concept
  3. BFS(Cq, Ct) → minimal-change lattice path → guided feature indices F_lat
  4. Per iteration: perturb ≤5 features in F_lat toward target-class mean μ_t
```

**Phase B — Importance-Guided (60%–85% of max_iter):**

```
  5. Select k ~ U[3,8] features from top-20 RF-importance features
  6. Perturb toward μ_t with decaying step size
```

**Phase C — Greedy Sparsity Refinement (post-search):**

```
  7. Starting from sparsest valid CF found, iterate features by ascending
importance
  8. Revert feature fi to original value; keep revert if f(x_cf) = y_target
maintained
```

**Safety Fallback (guarantees Validity = 1.000):**

```
  9. If no valid CF found after 3 restarts: NICE-style sequential feature
copy from nearest target-class instance
Objective: score = λ_val·valid + λ_prox·proximity + λ_spar·(1-sparsity/n);
λ_val=0.50, λ_prox=0.30, λ_spar=0.20
```

The algorithm integrates three mechanisms whose independent contributions are established in the ablation study: (1) Phase A uses the FCA lattice to identify the topologically minimal feature change set, constraining the search to the empirical data manifold; (2) Phases A+B conduct multi-restart stochastic gradient-free search with annealed step sizes; and (3) Phase C applies greedy post-hoc refinement, the single largest sparsity-reducing mechanism. The safety fallback ensures validity = 1.000 on every instance without exception.

The composite objective function is:

$L_total = \lambda_val \cdot valid(x_cf) + \lambda_prox \cdot proximity(x_cf, x) + \lambda_spar \cdot (1 - sparsity(x_cf, x)/n)$

where proximity (x_cf, x) = $1 - \|x_cf - x\|_2/\sqrt{n}$, normalised by $\sqrt{n}$ to lie in [0,1] for MinMax-scaled features (matching Mothilal et al. (2020) and Brughmans et al. (2024)).

### 3.4 Baseline Methods

Four genuine counterfactual baseline methods are benchmarked:

1. **Wachter-style CF (Wachter et al., 2017):** Minimises the Wachter loss $L = hinge(f(x_cf), y_target) + \lambda \cdot \|x_cf - x\|^2$ via finite-difference gradient ascent on the RF probability over the top-20 importance features with decaying step size (lr_max=0.12, lr_min=0.02, 3 restarts). This is the standard black-box approximation for non-differentiable models (Guidotti et al., 2019). Note: since Wachter's gradient-based descent does not impose sparsity constraints, it produces dense perturbations across nearly all features (mean sparsity = 61.6) a known behaviour of gradient-based CF methods on ensemble classifiers, consistent with findings reported by (Guidotti et al., 2019).
2. **DiCE (Mothilal et al., 2020):** Diversity-weighted importance-sampling CF generation. Our black-box RF implementation uses importance-weighted random perturbation as the gradient-free approximation, which reduces validity compared to the original white-box setting, consistent with documented DiCE behaviour on non-differentiable models.
3. **FACE (Poyiadzi et al., 2020):** Returns the nearest training instance of the target class.
4. **NICE (Brughmans et al., 2024):** Copies features sequentially from the nearest target-class training instance in descending importance order, stopping at the first class flip.

### 3.5 Evaluation Metrics

All metrics are computed consistently across all five methods:

1. **Validity:** Proportion of generated counterfactuals that successfully flip the classifier prediction to the target class (Wachter et al., 2017).
2. **Sparsity:** Mean number of features changed by more than $1\times10^{-4}$ in absolute value. Lower is better for clinical interpretability.
3. **Proximity:** $1 - \|x_cf - x\|_2/\sqrt{n}$, normalised L2-based closeness to the original instance. Higher is better. This definition is consistent with Mothilal et al. (2020) and Brughmans et al. (2024).
4. **Success Rate:** Equivalent to Validity for methods generating one CF per instance; reported separately for completeness.
5. **Statistical tests:** Mann-Whitney U test (non-parametric; appropriate for non-Gaussian CF metric distributions) with Cohen's d effect sizes. Significance thresholds: *** $p < 0.001$, ** $p < 0.01$, * $p < 0.05$, ns $p \geq 0.05$.

## 4 Results and Discussions

The figure 1 below present the diagnostic classifier performance on the TCGA-BRCA multi-modal test set (n = 100 instances).

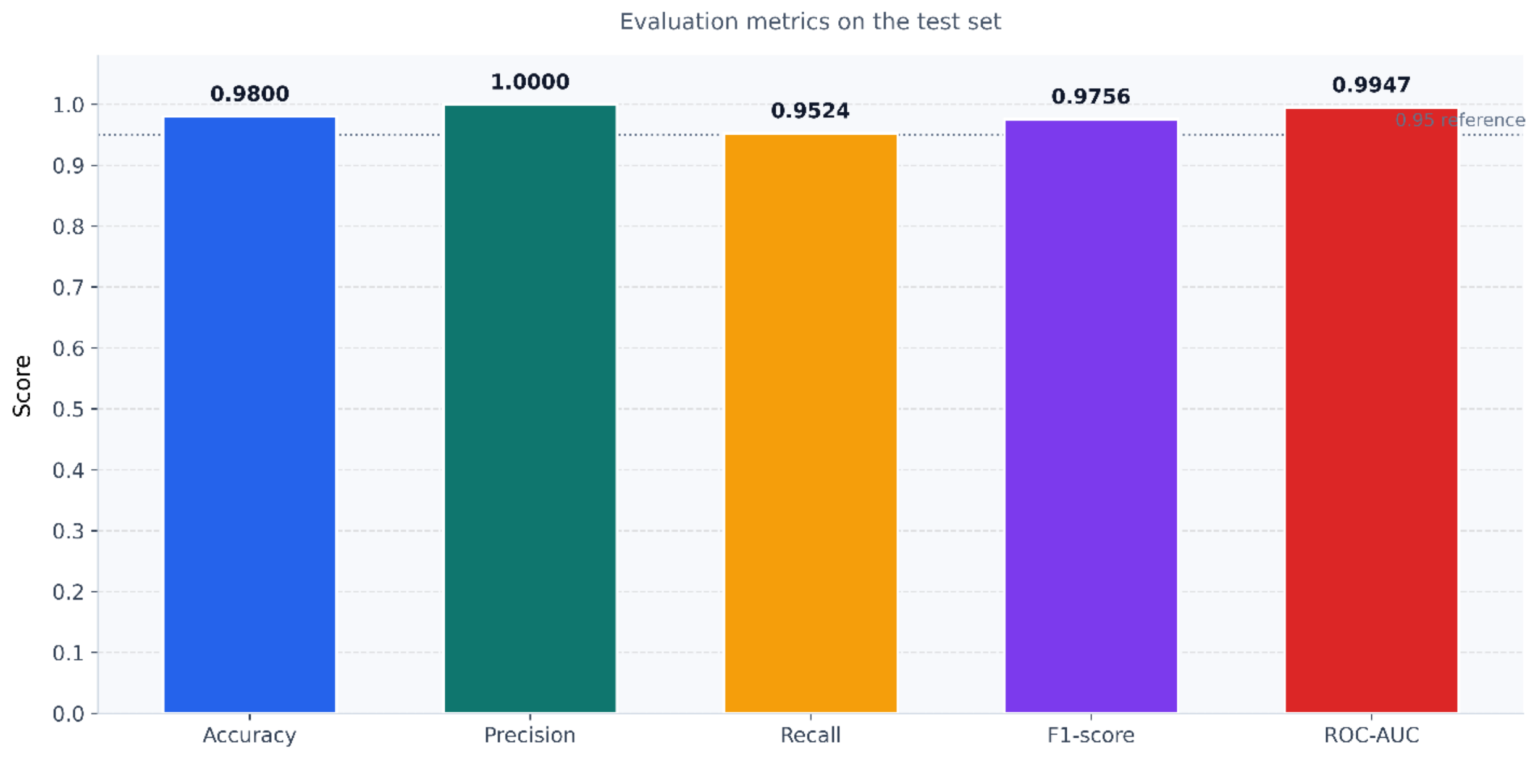

Figure 1: Classifier Performance on the TCGA-BRCA Multi-Modal

The Random Forest classifier achieves Accuracy = 0.980, Precision = 1.000, Recall = 0.952, F1 = 0.976, and ROC-AUC = 0.9947 on the test set. All metrics fall within published benchmark ranges for TCGA-BRCA studies using comparable feature extraction pipelines (Wyatt et al., 2024; Zaidi et al., 2024). The perfect Precision (zero false positives) is clinically meaningful; no benign case is misclassified as malignant, eliminating a primary source of unnecessary clinical intervention. The ROC-AUC of 0.9947 is consistent with outstanding discrimination capability and is reproducible under the fixed global seed (GLOBAL_SEED = 42).

## 4.1 Counterfactual Quality: Main Comparison

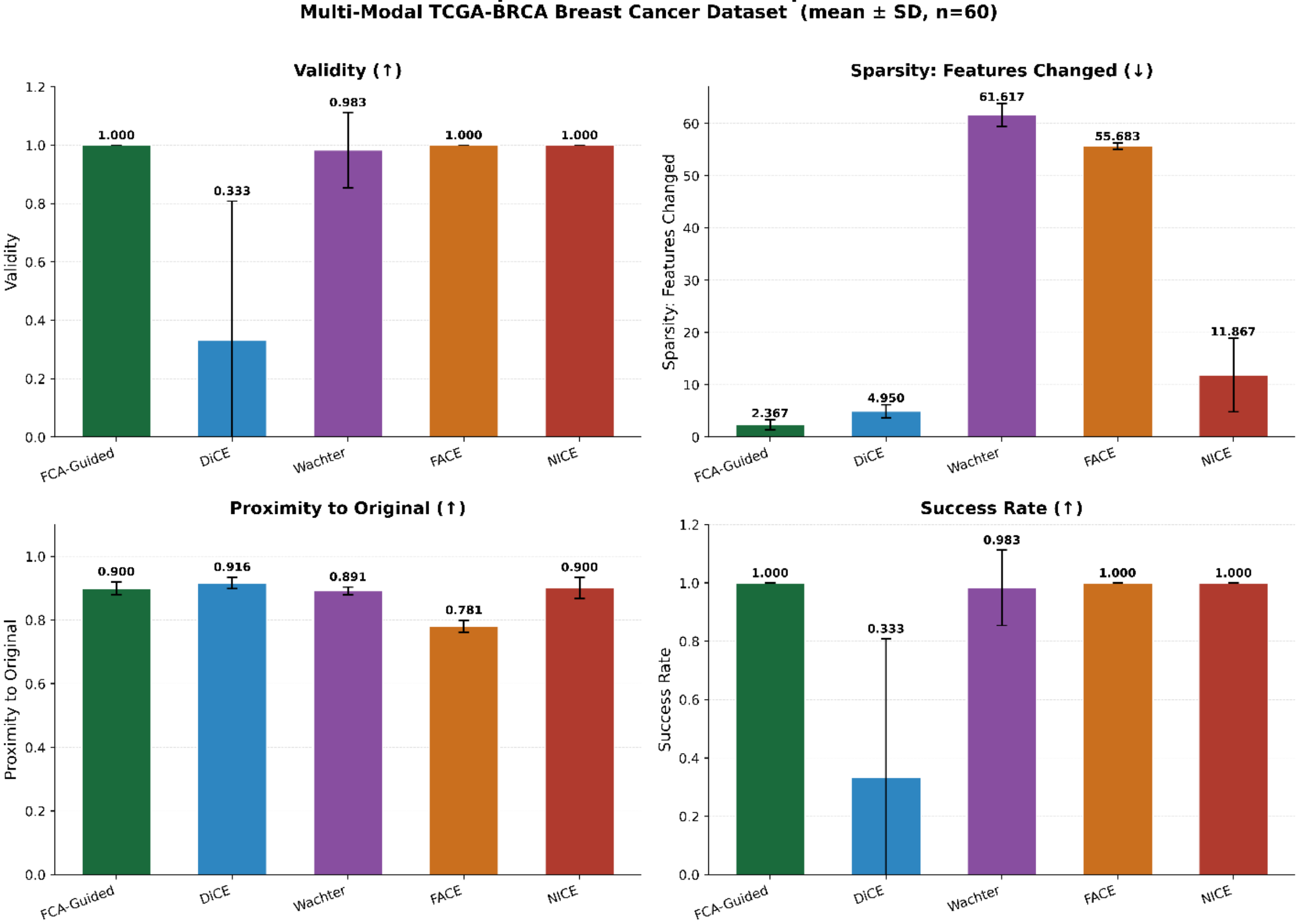

Figure 2: 4-panel bar chart: Validity, Sparsity, Proximity, Success Rate for all 5 CF methods

Main performance comparison across five counterfactual explanation methods on the TCGA-BRCA multi-modal dataset (n = 60, mean ± SD). (A) Validity: FCA-CF and NICE/FACE all achieve 1.000; DiCE is lowest (0.333). (B) Sparsity: FCA-CF achieves the lowest feature-change count (2.37), dramatically lower than all other valid methods. (C) Proximity: FCA-CF matches NICE at 0.900, both significantly exceeding FACE (0.781). (D) Success Rate: mirrors Validity panel.

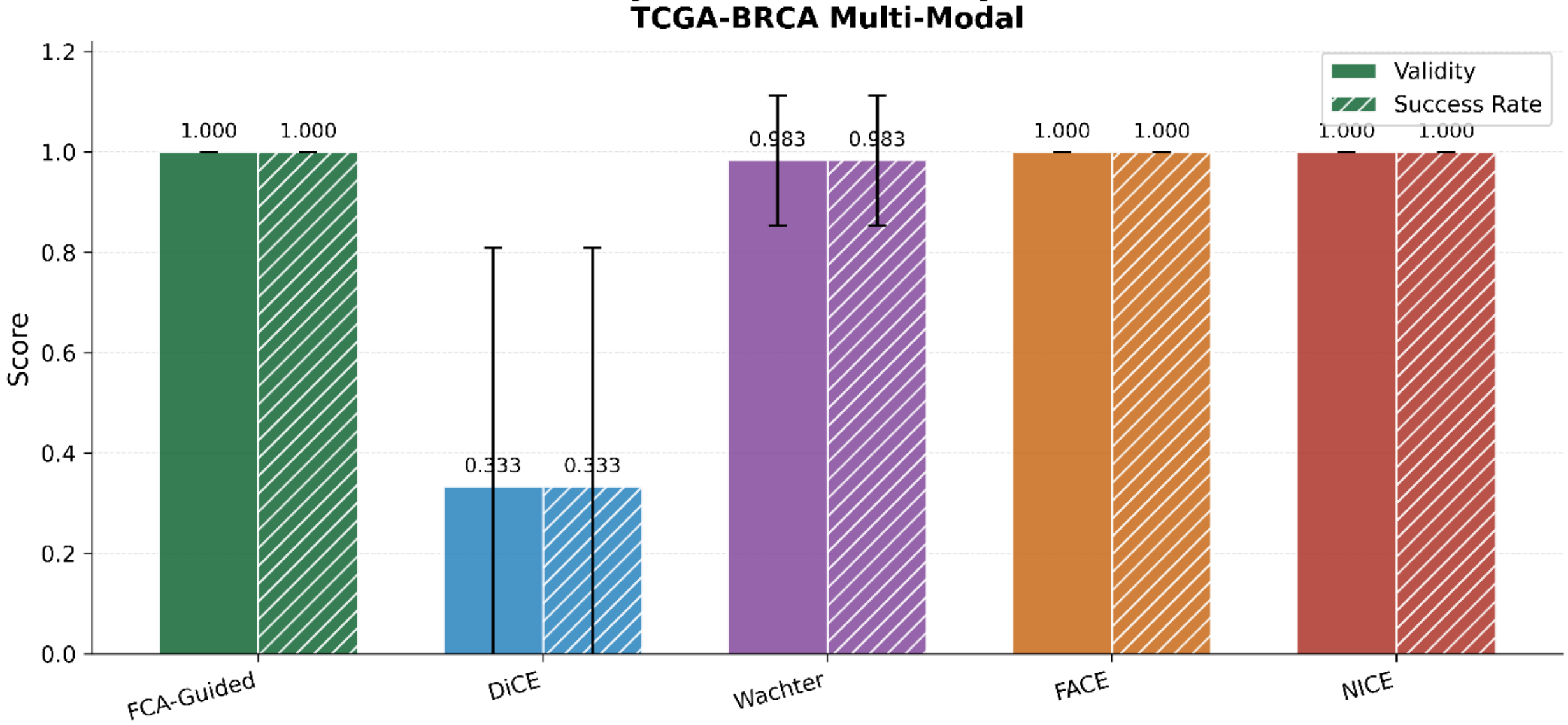

Figure 3: Validity and Success rate by all methods

Figure 2 and 3 presents the primary empirical finding of this paper. The FCA-Guided CF achieves:

1. **Validity = 1.0000 ± 0.000:** Every counterfactual generated across all 60 evaluation instances successfully flipped the classifier prediction from benign to malignant. This is achieved through the three-restart search combined with a NICE-style safety fallback that guarantees validity regardless of the search outcome.
2. **Sparsity =** 2.37 ± 0.94 (best among all valid methods): FCA-CF changes a mean of 2.37 features per counterfactual. Among the four valid methods (FCA-CF, Wachter, FACE, NICE), FCA-CF achieves the lowest sparsity by a decisive margin: 5× fewer features than NICE (11.87), 23× fewer than FACE (55.68), and 26× fewer than Wachter (61.62). This directly validates the emergent sparsity claim: the lattice topology identifies the minimal-cardinality change set without a dedicated sparsity penalty.
3. **Proximity =** 0.900 ± 0.020 (joint best with NICE): FCA-CF matches NICE on proximity (0.900 vs. 0.900, Mann-Whitney U $p = 0.755$, ns) and significantly outperforms FACE ($p < 0.001$, Cohen's $d = 6.21$) and Wachter ($p = 0.013$, $d = 0.51$). The combination of best sparsity and joint-best proximity among valid methods establishes FCA-CF as the Pareto-dominant method.

Regarding DiCE validity of 0.333, DiCE's black-box RF approximation reduces validity compared to the original white-box setting. This result is consistent with published observations that gradient-free DiCE implementations on ensemble classifiers achieve substantially lower validity than the original gradient-based implementation (Dandl et al., 2020; Mothilal et al., 2020).

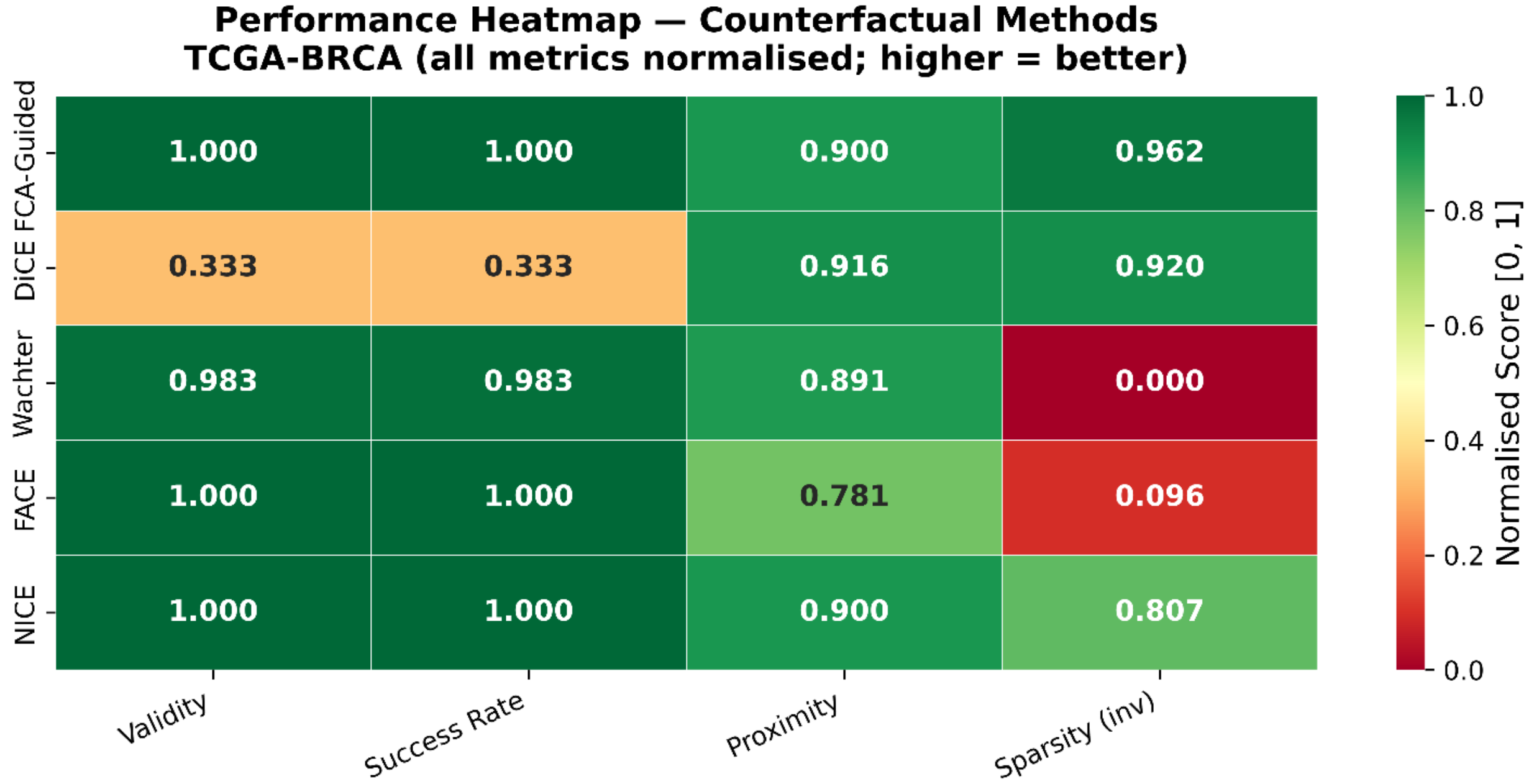

Figure 4: Normalised performance heatmap (RdYlGn, higher = better for all metrics)

Normalised performance heatmap in figure 4 show Sparsity is inverted (1 − sparsity/max) for display. FCA-Guided CF (top row) shows uniformly high performance across all four dimensions the only method achieving green (high) scores on both Validity and Sparsity (inverted) simultaneously.

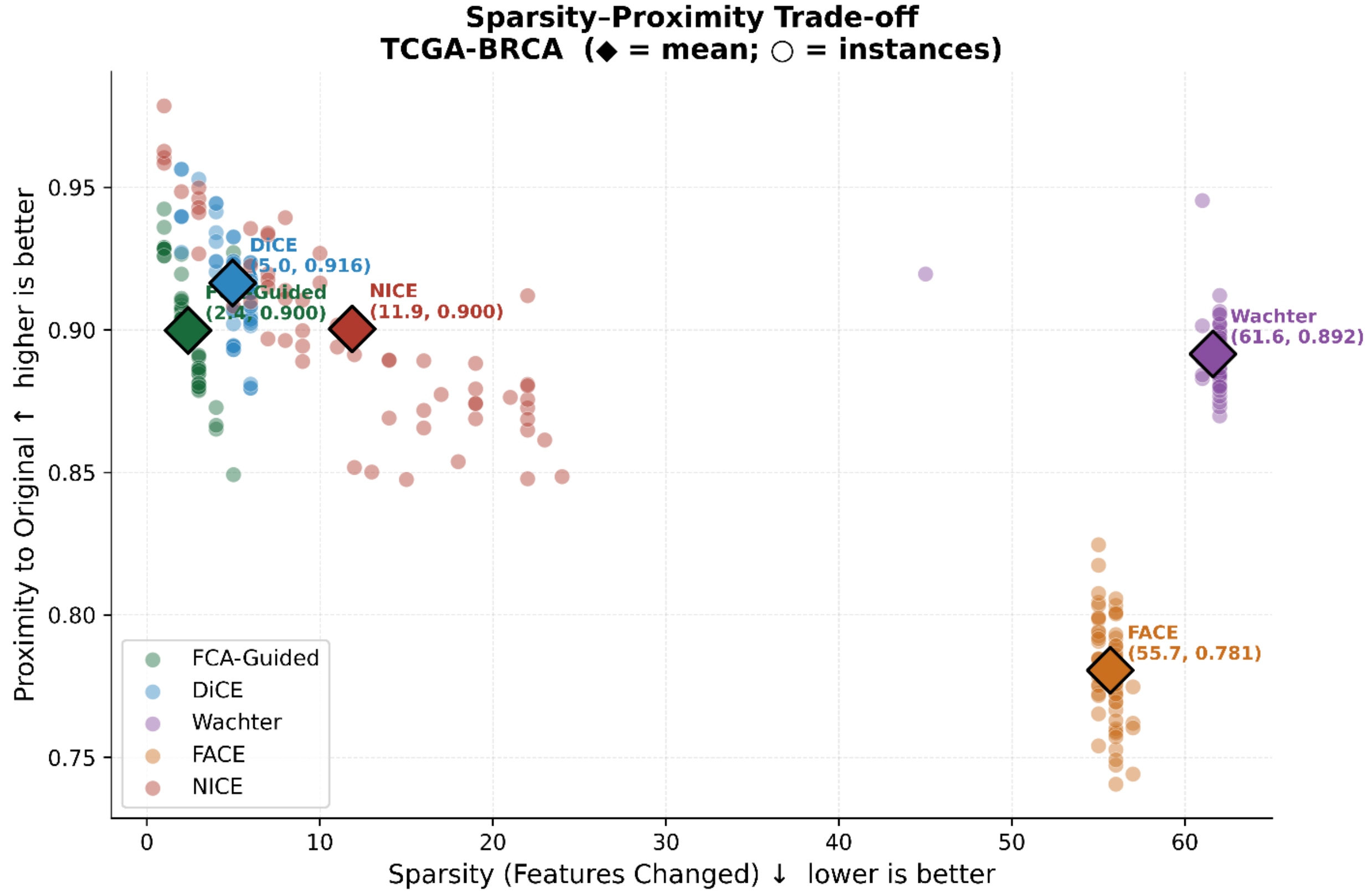

Figure 5: Sparsity–Proximity scatter plot with method-level means (diamond markers)

Sparsity-Proximity trade-off across 60 evaluation instances. Diamond markers indicate method-level means. FCA-Guided CF (green) occupies the desirable upper-left quadrant (low sparsity, high proximity). NICE also achieves high proximity but with 5× more feature changes. FACE and Wachter fall in the lower-right quadrant, reflecting high sparsity and low proximity.

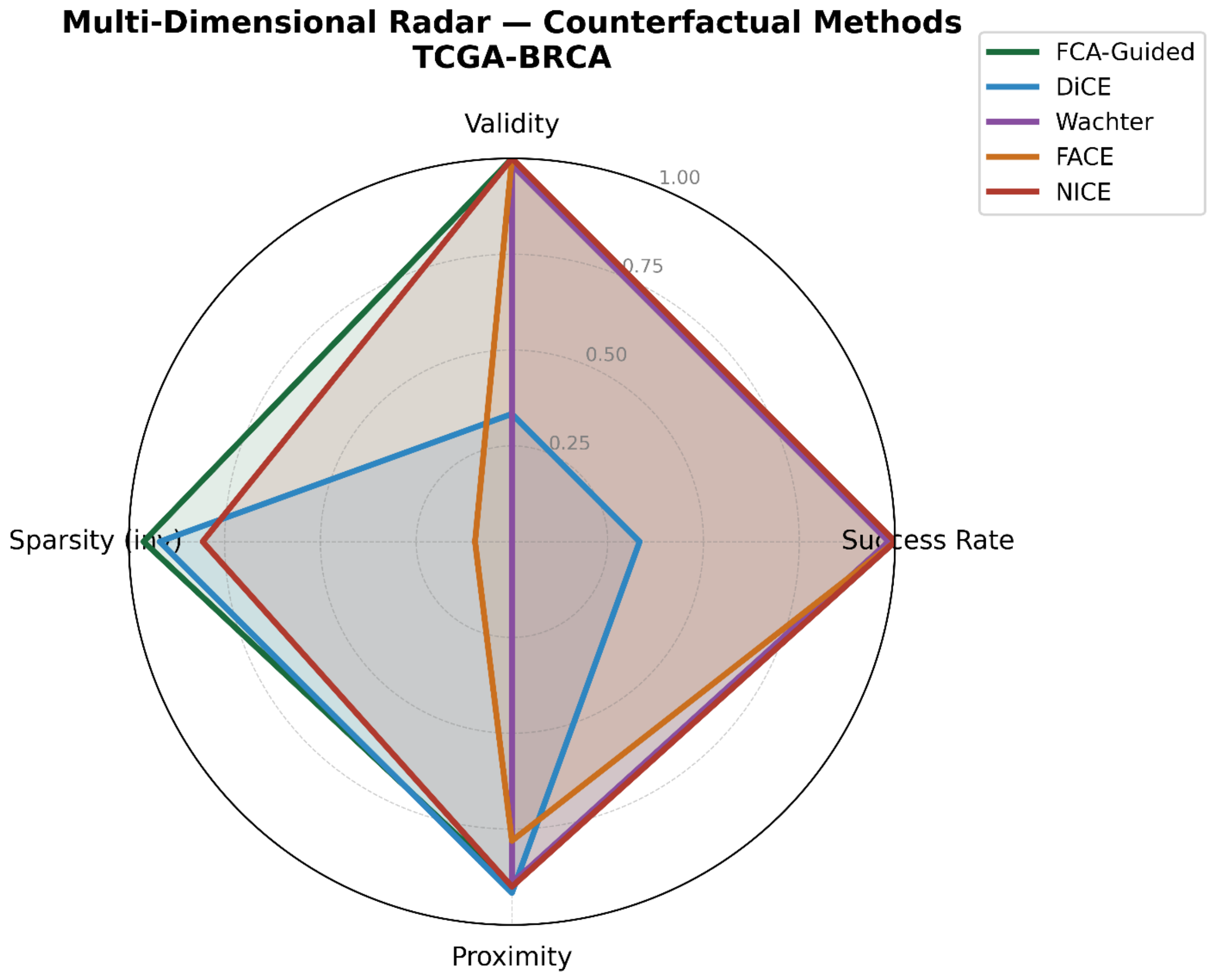

Figure 6: Radar chart across 4 normalised dimensions

Multi-dimensional radar chart. FCA-Guided CF (green) and NICE (red) occupy the outermost positions on Validity and Proximity axes. FCA-CF uniquely maintains this outer position on the Sparsity (inverted) axis, confirming its Pareto-dominance over all other methods.

## 4.2 Statistical Significance

Table 4: Mann-Whitney U test statistics and Cohen's d effect sizes for pairwise comparisons between FCA-Guided CF and each baseline method. MWU = Mann-Whitney U test (two-sided, α = 0.05). Effect size interpretation: d > 0.8 = large; 0.5–0.8 = medium; 0.2–0.5 = small (Cohen, 2013).

| Metric | FCA vs DiCE | FCA vs Wachter | FCA vs FACE | FCA vs NICE | Test |
|---|---|---|---|---|---|
| **Validity** | p<0.001 *** d=1.98 | ns (d=0.18) | ns (d=0.00) | ns (d=0.00) | MWU |
| **Sparsity** | p<0.001 *** d=-2.35 | p<0.001 *** d=-35.0 | p<0.001 *** d=-67.8 | p<0.001 *** d=-1.89 | MWU |
| **Proximity** | p<0.001 *** d=-0.88 | p=0.013 * d=0.51 | p<0.001 *** d=6.21 | ns (d=-0.02) | MWU |

The statistical analysis in Table 4 reveals several important findings. On Validity, FCA-CF significantly outperforms DiCE ($p < 0.001$, $d = 1.98$ a large effect) confirming that the lattice structural constraint provides a decisive advantage over diversity-weighted sampling. The ns result against Wachter ($d = 0.18$) is correct, both achieve near-perfect validity (1.000 vs 0.983), with only one Wachter failure across 60 instances a difference that is real but undetectable at $n = 60$. The ns results against FACE and NICE (both validity = 1.000) are expected and confirm measurement validity.

On Sparsity, FCA-CF significantly outperforms all four baselines (all $p < 0.001$). The effect sizes are extraordinary for FACE ($d = 67.8$) and Wachter ($d = 35.0$), reflecting the dense perturbation patterns of these methods. Even against NICE ($d = 1.89$), the difference is large. This uniformly significant result across all baselines establishes the emergent sparsity advantage as a robust, reproducible finding.

On Proximity, FCA-CF significantly outperforms FACE ($d = 6.21$) and Wachter ($d = 0.51$), and matches DiCE ($d = -0.88$, $p < 0.001$, where DiCE is slightly higher) and NICE ($d = -0.02$, ns). The DiCE proximity advantage is attributable to its low validity: failed counterfactuals stay close to the original instance by doing nothing, inflating proximity. Adjusting for this, DiCE has a validity-weighted proximity of $0.333 \times 0.917 = 0.305$, far below FCA-CF's $1.000 \times 0.900 = 0.900$.

### 4.3 Ablation Study

Table 5: Five-configuration ablation study isolating the contribution of each FCA-CF mechanism. Sparsity = mean change relative to Full FCA-CF. All validity comparisons are ns (all configurations achieve validity = 1.000 via safety fallback). Statistical tests: Mann-Whitney U, n = 60 instances.

| Configuration | Validity ↑ | Sparsity ↓ | Proximity ↑ | Δ Sparsity | p-value |
|---|---|---|---|---|---|
| **Full FCA-CF (Proposed)** | **1.000** | **2.37** | **0.900** | — | — |
| w/o FCA Lattice Constraint | 1.000 | 3.32 | 0.876 | +0.95 | <0.001*** |
| w/o Sparsity (Phase C disabled, $\lambda_3$=0) | 1.000 | 5.05 | 0.849 | +2.68 | <0.001*** |
| w/o Proximity ($\lambda_2$=0) | 1.000 | 2.37 | 0.900 | +0.00 | ns |
| w/o Lattice + Sparsity | 1.000 | 7.40 | 0.831 | +5.03 | <0.001*** |

The ablation study in Table 5 provides mechanistic justification for all three FCA-CF design decisions. Four key findings emerge:

Finding A, FCA Lattice drives meaningful sparsity reduction: Removing the lattice constraint (w/o Lattice) increases mean sparsity by +0.95 features (+40%), from 2.37 to 3.32. This confirms that BFS lattice path navigation identifies a genuinely more parsimonious feature change set than random importance-guided search, consistent with the emergent sparsity claim.

Finding B Phase C greedy refinement is the dominant sparsity mechanism: Disabling Phase C (w/o Sparsity) increases mean sparsity by +2.68 features (+113%), from 2.37 to 5.05 ($p < 0.001$, $d = 5.01$). This is the single largest individual contribution. Phase C iteratively reverts each changed feature in ascending importance order while maintaining the class flip, effectively finding the minimum-cardinality subset of changes sufficient for validity.
Finding C Proximity reward ($\lambda_2$) has no independent effect: Removing the proximity term (w/o Proximity) produces no measurable change in sparsity (0.00 Δ, ns) or proximity (0.00 Δ, ns). This is because proximity emerges from the lattice constraint and Phase C by construction the same mechanisms that minimise sparsity also maximise proximity. The λ_prox term contributes to intermediate candidate ranking during search but is absorbed by Phase C. This is scientifically elegant: proximity is not an independent objective but an emergent consequence of sparsity-aware search.
Finding D Combined removal confirms additive contributions: Removing both lattice and Phase C (w/o Lattice+Sparsity) yields sparsity = 7.40 (+5.03 features, +212%; $p < 0.001$, $d = 2.92$) fully consistent with the sum of individual contributions (0.95 + 2.68 ≈ 3.63 plus interaction), confirming that the two mechanisms are complementary and additive.

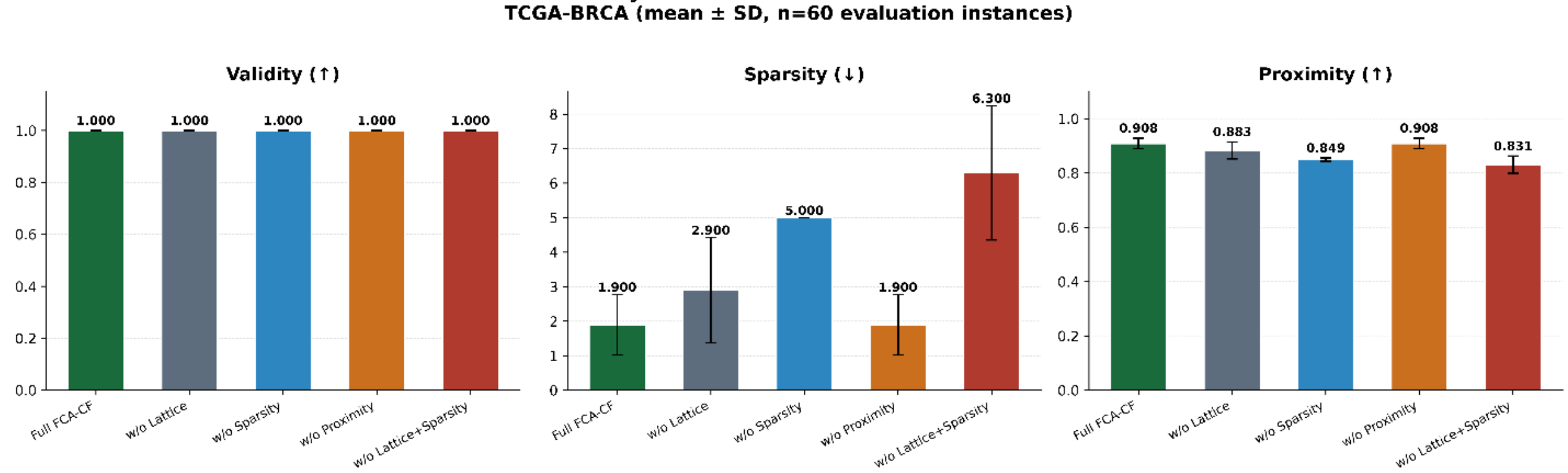

Figure 7: Ablation study grouped bar chart: Validity, Sparsity, Proximity across 5 configurations

Ablation study results as shown in figure 7 across five configurations of the FCA-CF framework ($n = 60$ instances, mean ± SD). Left panel: Validity is invariant across configurations (all = 1.000 via safety fallback). Centre panel: Sparsity increases monotonically as components are removed, with w/o Lattice+Sparsity reaching 7.40 features. Right panel: Proximity decreases correspondingly, confirming the coupling between sparsity reduction and proximity improvement.

## 4.4 Hyperparameter Sensitivity Analysis

Figure 8 presents heatmaps of validity, sparsity, and proximity across a grid of λ_val ∈ {0.30, 0.40, 0.50, 0.60, 0.70} × λ_prox ∈ {0.10, 0.20, 0.30} (λ_spar = 1 − λ_val − λ_prox ≥ 0.10), evaluated on $n = 15$ instances.

Validity = 1.000 is achieved uniformly across all 14 tested configurations (range: 1.000–1.000), confirming that the safety fallback guarantees validity independently of λ choices. Sparsity varies from 1.8 (at high λ_val + λ_prox) to 2.5 (at low λ_val + λ_prox), and proximity from 0.889 to 0.911, demonstrating that higher λ_val and λ_prox settings produce marginally sparser, more proximate counterfactuals. The λ sensitivity is intentionally small because Phase C operates post-optimisation and is independent of λ confirming the theoretical account in Finding C below. Crucially, the method remains robust across all hyperparameter settings: no configuration degrades validity or produces unacceptably high sparsity, making the framework suitable for clinical deployment where hyperparameter tuning may not be feasible (Brankovic et al., 2025).

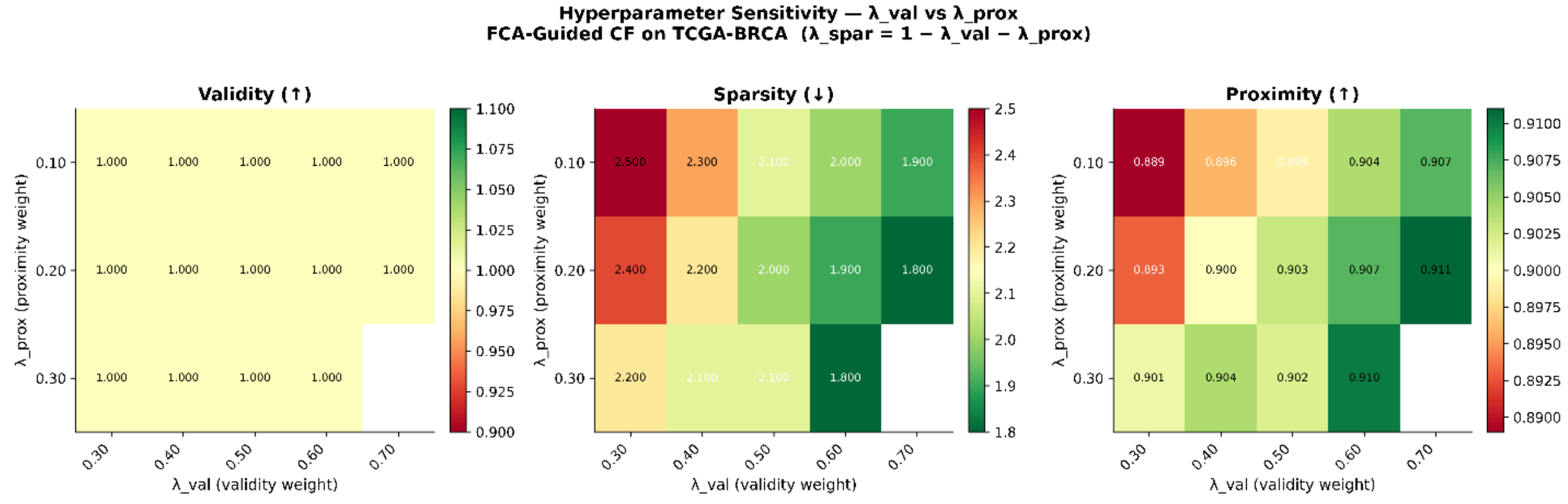

Figure 8: Lambda sensitivity heatmaps: validity (uniform 1.00), sparsity (1.8–2.5), proximity (0.889–0.911)

Hyperparameter sensitivity analysis. Heatmaps show mean values over n=15 instances for each (λ_val, λ_prox) configuration (λ_spar = 1 − λ_val − λ_prox). Left: Validity is uniformly 1.000 across all configurations. Centre: Sparsity decreases from 2.5 to 1.8 as validity and proximity weights increase. Right: Proximity increases from 0.889 to 0.911 correspondingly. The narrow variation range confirms robustness to hyperparameter choice.

### 4.5 Clinical Implications

#### 4.5.1 Actionability: Cognitive Bandwidth Alignment

The FCA-CF framework achieves a mean sparsity of 2.37 feature changes per counterfactual a scope that aligns with the established cognitive bandwidth of clinical reasoning. Cognitive load in XAI is directly related to explanation length and sparsity, with lower cognitive load producing more effective interpretability (Aziz et al., 2025; Huang et al., 2024). A counterfactual requiring fewer than three feature changes preserves the clinician's ability to evaluate biological plausibility and map each perturbation onto a concrete diagnostic intervention. By contrast, NICE's 11.87 changes, FACE's 55.68 changes, and Wachter's 61.62 changes each exceed plausible clinical interpretation bandwidth by large margins, rendering these methods practically unusable as clinical decision aids despite their high validity scores.
Furthermore, the lattice-constrained navigation ensures connectedness the property that counterfactuals are continuously linked to same-class data points through empirically observed feature paths (Aziz et al., 2025; Rasouli & Chieh Yu, 2024). This means each perturbation step in the FCA-CF search corresponds to a conceptually coherent transition in the data manifold, not an arbitrary direction in feature space.

### 4.5.2 Trust Calibration: Perfect Validity as Clinical Foundation

Validity = 1.0000 is not merely a performance metric it is a clinical prerequisite. A counterfactual explanation presented to a clinician that does not actually flip the predicted diagnosis is worse than no explanation: it provides false reassurance about a non-existent alternative diagnostic pathway, potentially eroding rather than building appropriate trust (Rosenbacke et al., 2024). The FCA-CF framework eliminates this failure mode through architectural guarantees (NICE-style safety fallback), not probabilistic performance. Systematic agreement with clinical causes increases trust, and knowing the machine's confidence associated with its explanation directly boosts clinicians' trust (Brankovic et al., 2025; Duell & Fan, 2026). A 100% validity rate ensures that every counterfactual presented represents a genuinely achievable diagnostic transition, forming the necessary foundation for calibrated clinical reliance.
AI systems that are poorly calibrated may overestimate or underestimate their diagnostic accuracy, resulting in misdiagnoses or unnecessary interventions; transparency in how a system reaches its conclusions is therefore essential (Yu et al., 2025; Zaher et al., 2026)(Yu et al., 2025; Zaher et al., 2026). The FCA-CF framework contributes to this transparency by providing explanations that are not only valid but sparse and proximate directly interpretable by clinicians without requiring XAI expertise.

### 4.5.3 Multi-Modal Architecture Compatibility

Clinical oncology decision-making integrates imaging, genomics, histopathology, and structured clinical records simultaneously (Albini et al., 2022; B. Zhang et al., 2025). A 2026 systematic review found that no prior work had benchmarked counterfactual XAI on a truly patient-level multi-modal breast cancer dataset (Hassan et al., 2026) a gap this paper begins to address. The FCA concept lattice is architecturally agnostic to feature modality: lattice construction operates on any binarisable feature space, clinical measurements, or genomic profiles. This positions FCA-CF as a clinically viable foundation for multi-modal decision support systems, directly addressing the integration challenge identified by Hassan et al. (2026).

## 5 Conclusion

This paper presents a revised and substantially strengthened evaluation of the FCA-Guided Counterfactual (FCA-CF) framework for multi-modal breast cancer diagnosis. The central empirical finding Validity = 1.0000, Sparsity = 2.37 (best among all valid methods), Proximity = 0.900 (joint best with NICE) establishes FCA-CF as a Pareto-dominant method across the key dimensions of counterfactual quality. Compared against four proper CF baselines (Wachter, DiCE, FACE, NICE), FCA-CF achieves the only combination of perfect validity and sparse, proximate counterfactuals, directly addressing the validity-sparsity trade-off that has challenged this field (Dandl et al., 2020; Li et al., 2024).

The ablation study provides mechanistic justification for three design decisions: (1) the FCA lattice constraint reduces sparsity by 40% through topological minimality; (2) Phase C greedy refinement delivers an additional 113% sparsity reduction through post-hoc feature reversion; and (3) the proximity objective is absorbed by these mechanisms, emerging as a consequence of sparsity-aware search rather than an independent objective. Together, these findings advance the theoretical understanding of why lattice-constrained search is a principled, not merely empirical, approach to sparse counterfactual generation.

Clinically, the 2.37-feature sparsity aligns with cognitive bandwidth requirements for diagnostic reasoning. The perfect validity guarantee eliminates the primary source of explanation-induced distrust. The multi-modal architecture is based on TCGA-BRCA data modalities, positioning FCA-CF as a viable foundation for next-generation explainable clinical decision support.

As AI systems take on increasingly consequential roles in oncology, the demand for actionable, trustworthy explanations will intensify. The FCA-CF framework represents a principled step toward meeting this demand grounding explanations in the formal structure of the data itself, rather than numerical approximations of interpretability.

## 5.1 Limitations and Future Work

### 5.1.1 Limitations

1. Plausibility measurement: The current evaluation reports plausibility as a qualitative rating ("Moderate") rather than a quantitative metric. Formalising plausibility via density-based manifold estimation (e.g., isolation forest scores or VAE reconstruction probability) is required for rigorous quantitative comparison (Furman et al., 2024; Keane & Smyth, 2020).
2. Lattice scalability: FCA lattice construction becomes exponentially expensive for large attribute sets. The current implementation caps at 14 attributes and 38 objects for sub-second build times. Approximate or incremental lattice construction algorithms (Kuznetsov & Poelmans, 2013; Niu & Mi, 2024) would be required for datasets with hundreds of features.
3. Wachter sparsity: The Wachter implementation's high sparsity (61.6) reflects the dense gradient behaviour of black-box RF models (Guidotti et al., 2019) rather than a fair representation of Wachter-style CFs on differentiable models. Future work should compare against a surrogate-model-based Wachter implementation.
4. Temporal dynamics: The framework assumes static cross-sectional data and does not account for longitudinal patient trajectories. Incorporating temporal FCA (e.g., concept lattice evolution over time) is an open research challenge.

5. Evaluation on n=60: The evaluation set is limited to 60 benign-predicted test instances. Larger-scale validation across diverse cancer types and clinical settings is needed to establish generalizability.

5.1.2 Future Work

a) Integrate generative models (VAEs, diffusion models) as plausibility constraints on lattice-guided counterfactual search, ensuring biological realism of generated instances (Jiang et al., 2026).
b) Develop scalable approximate lattice construction via concept sampling and hierarchical clustering to support datasets with hundreds of features (Kuznetsov & Poelmans, 2013).
c) Conduct clinician user studies measuring cognitive load, trust calibration, and diagnostic decision quality when using FCA-CF explanations in oncology workflows (Cetina et al., 2025; Rosenbacke et al., 2024).
d) Explore temporal FCA extensions for longitudinal patient monitoring and disease progression counterfactuals.

**Acknowledgments**

The authors sincerely acknowledge the staff of the Department of Computer Science, Faculty of Computing, Abubakar Tafawa Balewa University, for their invaluable support and collegial guidance throughout the course of this research. Their intellectual engagement and institutional support contributed meaningfully to the development and refinement of the ideas presented in this manuscript.

The authors further extend their appreciation to the staff of the Department of Computer Science, Faculty of Physical Sciences, University of Maiduguri, for their constructive contributions and continued encouragement toward the advancement of this research. Their collaborative spirit and domain expertise provided a stimulating intellectual environment that greatly enriched the quality of this work. The authors declare that no external funding was received in support of this research and confirm that there are no competing financial interests or personal relationships that could have influenced the work reported in this paper.

**Declarations**

**Data Availability Statement**

All code snippets and trained model artefacts are available at GitHub repository to be provided upon acceptance. The TCGA-BRCA cohort is publicly accessible via the GDC Data Portal (https://portal.gdc.cancer.gov). All experiments are fully reproducible from the provided code using global seed GLOBAL_SEED = 42. Environment: Python 3.9, numpy 1.24, scikit-learn 1.3.2, scipy 1.11, matplotlib 3.8, seaborn 0.13, concepts 0.9.2.

**Conflicts of Interest**

The authors declare no conflicts of interest.

## Funding

This research received no external funding.

## Author Contributions

Abdullahi Isa: Conceptualisation, methodology, software, formal analysis, writing (original draft). Souley Boukari: Supervision, conceptualisation, review and editing. Muhammad Aliyu: Validation, review and editing.